\documentclass[10pt,twocolumn,letterpaper]{article}

\pdfoutput=1

\usepackage{iccv}
\usepackage{times}
\usepackage{epsfig}
\usepackage{graphicx}     
\usepackage{amsmath}
\usepackage{amssymb}
\usepackage{booktabs}
\usepackage{multirow}     %
\usepackage{subcaption}   %
\usepackage{arydshln}     %
\usepackage[font=small,skip=0pt]{caption}
\usepackage{chngcntr}
\usepackage[dvipsnames]{xcolor}

\usepackage[pagebackref=true,breaklinks=true,letterpaper=true,colorlinks,bookmarks=false]{hyperref}

\iccvfinalcopy %

\ificcvfinal\fi

\newcommand\blfootnote[1]{%
	\begingroup
	\renewcommand\thefootnote{}\footnote{#1}%
	\addtocounter{footnote}{-1}%
	\endgroup
}

\begin{document}

\title{Improved Conditional VRNNs for Video Prediction}

\author{Llu\'is Castrej\'on$^{\dagger, \ddagger}$ \hspace{4em} Nicolas Ballas$^{\ddagger}$ \hspace{4em} Aaron Courville$^{\dagger, \P}$\\
$\dagger$ Mila - Universit\'e de Montr\'eal \hspace{3em} $\ddagger$ Facebook AI Research \\
$\P$ Canadian Institute for Advanced Research (CIFAR)\\
}

\maketitle

\begin{abstract}

Predicting future frames for a video sequence is a challenging generative modeling task.
Promising approaches include probabilistic latent variable models such as the Variational Auto-Encoder.
While VAEs can handle uncertainty and model multiple possible future outcomes, they have a tendency to produce blurry predictions.
In this work we argue that this is a sign of underfitting. To address this issue, we propose to increase the expressiveness of the latent distributions and to use higher capacity likelihood models.
Our approach relies on a hierarchy of latent variables, which defines a family of flexible prior and posterior distributions in order to better model the probability of future sequences.
We validate our proposal through a series of ablation experiments and compare our approach to current state-of-the-art latent variable models. Our method performs favorably under several metrics in three different datasets.

\end{abstract}

\section{Introduction}

\blfootnote{Correspondence to \textit{lluis.enric.castrejon.subira@umontreal.ca}}We investigate the task of video prediction, a particular instantiation of self-supervision~\cite{devlin2018bert, gidaris2018unsupervised} where generative models learn to predict future frames in a video.
Training such models does not require any annotated data, yet the models need to capture a notion of the complex dynamics of real-world phenomena (such as physical interactions) to generate coherent sequences.

\begingroup
\setlength{\tabcolsep}{0.5pt} %
\renewcommand{\arraystretch}{1} %
\begin{figure}
    \small
    \centering
    \begin{tabular}{cc|cccc}
            \multicolumn{2}{c|}{\textbf{Context}}  &
            \multicolumn{4}{|c}{\textbf{Predicted Frames}} 
            \\
            &
            $t = 2$ &
            $t = 3$ & 
            $t = 5$ & 
            $t = 10$ &
            $t = 20$ \\
            \hline
            {\rotatebox[origin=c]{90}{GT}} &
            \raisebox{-.4\height}{\includegraphics[scale=0.35]{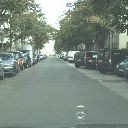}} &
            \raisebox{-.4\height}{\includegraphics[scale=0.35]{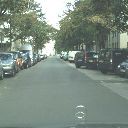}} & 
            \raisebox{-.4\height}{\includegraphics[scale=0.35]{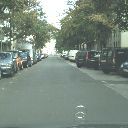}} &
            \raisebox{-.4\height}{\includegraphics[scale=0.35]{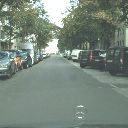}} &
            \raisebox{-.4\height}{\includegraphics[scale=0.35]{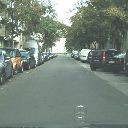}}
            \\[15.5pt]
            \multicolumn{2}{c|}{\textsc{SVG-LP \newline ~\cite{svg}}} &
            \raisebox{-.4\height}{\includegraphics[scale=0.35]{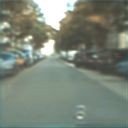}} & 
            \raisebox{-.4\height}{\includegraphics[scale=0.35]{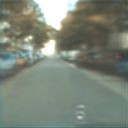}} &
            \raisebox{-.4\height}{\includegraphics[scale=0.35]{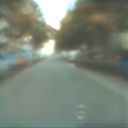}} &
            \raisebox{-.4\height}{\includegraphics[scale=0.35]{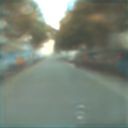}} 
            \\[15.5pt]
            \multicolumn{2}{c|}{\textsc{Ours}} &
            \raisebox{-.4\height}{\includegraphics[scale=0.35]{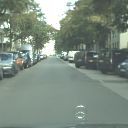}} & 
            \raisebox{-.4\height}{\includegraphics[scale=0.35]{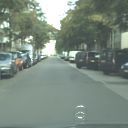}} &
            \raisebox{-.4\height}{\includegraphics[scale=0.35]{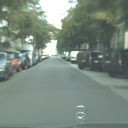}} &
            \raisebox{-.4\height}{\includegraphics[scale=0.35]{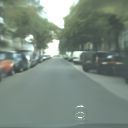}} 
    \end{tabular}
    \caption{
    \textbf{Can generative models predict the future?}
    We propose an improved VAE model for video prediction.
    Our model uses hierarchical latents and a higher capacity likelihood network to improve upon previous VAE approaches, generating more visually appealing samples that remain coherent for longer temporal horizons.
    }
    \label{fig:teaser}
\end{figure}
\endgroup

Uncertainty is an inherent difficulty associated with video prediction, as many future outcomes are plausible for a given sequence of observations~\cite{sv2p, svg}.
Predictions from deterministic models rapidly degrade over time as uncertainty grows, converging to an average of the possible future outcomes~\cite{video_lstm}. 
To address this issue, probabilistic latent variable models such as Variational Auto-Encoders (VAEs)~\cite{kingma2013auto, rezende2014stochastic}, and more specifically Variational Recurrent Neural Networks (VRNNs)~\cite{vrnn}, have been proposed for video prediction~\cite{sv2p, svg}. 
These models define a prior distribution over a set of latent variables, allowing different samples from these latents to capture multiple outcomes.

It has been empirically observed that VAE and VRNN-based models produce blurry predictions~\cite{larsen2015autoencoding,savp}.
This tendency is usually attributed to the use of a similarity metric in pixel space~\cite{larsen2015autoencoding,mathieu2015deep} such as Mean Squared Error (corresponding to a log-likelihood loss under a fully factorized Gaussian distribution). This has lead to alternative models such as VAE-GAN \cite{larsen2015autoencoding, savp}, which extends the traditional VAE objective with an adversarial loss in order to obtain more visually compelling generations.

In addition, the lack of expressive latent distributions has been shown to lead to poor model fitting~\cite{hoffman2016elbo}.
Training VAEs involves defining an approximate posterior distribution over the latent variables which models their probability after the generated data has been observed.
If the approximate posterior is too constrained, it will not be able to match the true posterior distribution and this will prevent the model from accurately fitting the training data.
On the other hand, the prior distribution over the latent variables can be interpreted as a model of uncertainty.

The decoder or likelihood network needs to transform latent samples into data observations covering all plausible outcomes. Given a simple prior, this transformation can be very complex and require high capacity networks.
We hypothesize that the reduced expressiveness of current VRNN models is limiting the quality of their predictions and investigate two main directions to improve video prediction models. 
First, we propose to scale the capacity of the likelihood network. We empirically demonstrate that by using a high capacity decoder we can ease the latent modeling problem and better fit the data. %

Second, we introduce more flexible posterior and prior distributions~\cite{sonderby2016train}.
Current video prediction models usually rely on one shallow level of latent variables and the prior and approximate posterior are parameterized using diagonal Gaussian distributions~\cite{sv2p}. %
We extend the VRNN formulation by proposing a hierarchical variant that uses multiple levels of latents per timestep.

Models leveraging a hierarchy of latents are known to be hard to optimize as they are required to backpropagate through a stack of stochastic latent variables, usually resulting in models that only make use of a small subset of the latents~\cite{kingma2013auto, maaloe2016auxiliary,sonderby2016train}. 
We mitigate this problem by using a warmup regime for the KL loss~\cite{sonderby2016ladder} and a dense connectivity pattern~\cite{huang2017densely, maaloe2019biva} between the input and latent variables. 
Specifically, each stochastic latent variable is connected to  the input and to all subsequent stochastic levels in the hierarchy. 
Our empirical findings confirm that only with these techniques our model is able to take advantage of different layers in a latent hierarchy. 

We validate our hierarchical VRNN in three datasets with varying levels of future uncertainty and realism: Stochastic Moving MNIST ~\cite{svg}, the BAIR Push Dataset ~\cite{dataset_bair_pushing} and Cityscapes ~\cite{dataset_cityscapes}. When compared to current state of the art models~\cite{svg, savp}, our approach performs favorably under several metrics. %
In particular for the BAIR Push Dataset, our hierarchical-VRNN shows an improvement of $44\%$ in Video Fr\'echet Distance (FVD)~\cite{fvd} and $9.8\%$ in term of LPIPS score~\cite{lpips} over SVG-LP~\cite{svg}, the previous best VAE-based model. It also achieves a similar FVD than the SAVP VAE-GAN model~\cite{savp}, while showing a $11.2\%$ improvement in terms of LPIPS over this baseline.

\section{Related Work}

Initial video prediction approaches relied on deterministic models. Ranzato et al.~\cite{ranzato2014video} divided frames into patches and predicted their evolution in time given previous neighboring patches.
In~\cite{video_lstm} Srivastava et al.~used LSTM networks on pretrained image embeddings to predict the future.
Similarly, Oh et al.~\cite{oh2015action} used LSTMs on CNN representations to predict frames in Atari games when given the player actions. 

ConvLSTMs~\cite{xingjian2015convolutional} adapt the LSTM equations to spatial feature maps by replacing matrix multiplications with convolutions. They were originally used for precipitation nowcasting and are commonly used for video prediction. 

Other works have proposed to disentangle the motion and context of the frames to generate~\cite{villegas2017decomposing, tulyakov2018mocogan, denton2017unsupervised}. They assume that a scene can be decomposed as multiple objects,
which allows them to use a fixed representation for the background.
Our approach does not follow this modeling assumption and instead tries to capture the uncertainty in the future.

Autoregressive models~\cite{kalchbrenner2017video, reed2017parallel} approximate the full joint data distribution $p(x_1, x_2, ..., x_N)$ over pixels, which allows them to capture complex pixel dependencies but at the expense of making their inference mechanism slow and not scalable to high resolutions. 
Latent variable models using GANs~\cite{goodfellow2014generative} were proposed in ~\cite{vondrick2016generating, vondrick2016anticipating, tulyakov2018mocogan}. Training pure GAN video models is still an open research direction: training is unstable and most models require auxiliary losses. 

A successful approach so far has been based on VAE~\cite{kingma2013auto, rezende2014stochastic}/VRNN~\cite{vrnn} models. 
SV2P~\cite{sv2p} proposed to capture sequence uncertainty in a single set of latent variables kept fixed for each predicted sequence. 
SVG~\cite{svg} adopted the VRNN formulation~\cite{vrnn}, introducing per-step latent variables (SVG-FP) and a variant with a learned prior (SVG-LP), which makes the prior at a certain timestep a function of previous frames. 
In recent work, SAVP~\cite{savp} proposed to use the VAE-GAN~\cite{larsen2015autoencoding} framework for video, a hybrid model that offers a trade-off between VAEs and GANs. 
Our model extends the VRNN formulation by introducing a hierarchy of latents to better approximate the data likelihood.

There are multiple works addressing hierarchical VAEs for non-sequential data~\cite{ranganath2016hierarchical,maaloe2016auxiliary, sonderby2016ladder, kingma2016improved}. 
While hierarchical VAEs can model more flexible latent distributions, training them is usually difficult due to the multiple layers of conditional latents~\cite{sonderby2016train}. %
Ladder Variational Autoencoders~\cite{sonderby2016ladder} proposed a series of techniques to partially alleviate this issue. IAF~\cite{kingma2016improved} used a similar architecture to Ladder VAEs and extended it with a novel normalizing flow. Recent work~\cite{maaloe2019biva} has trained very deep hierarchical models that produce visually compelling samples. We extend hierarchical latent variable models to sequential data and apply them to video prediction. 
Concurrent work ~\cite{kumar2019videoflow} has proposed a fully invertible model for video.

\section{Preliminaries}

We follow previous work in video prediction~\cite{svg}. 
Given  $D$ context frames $\mathbf{c} = (c_1, c_2, ..., c_D)$ and the $T$ following future frames $\mathbf{x} = (x_1, x_2, ..., x_T)$, our goal is to learn a generative model that maximizes the probability $p(\mathbf{x}|\mathbf{c})$.

VRNN follows the VAE formalism and introduces a set of latent variables $\mathbf{z}= (z_1, z_2, ..., z_T)$ to capture the variations in the observed variables $\mathbf{x}$ at each timestep $t$. 
It defines a \textit{likelihood} model $p(\mathbf{x}|\mathbf{z}, \mathbf{c}) = \prod_{t=1}^T  p(x_t | \mathbf{z_{\leq t}}, \mathbf{x_{<t}}, \mathbf{c})$ and a \textit{prior} distribution $p(\mathbf{z}| \mathbf{c}) = \prod_{t=1}^T  p(z_t| \mathbf{z_{<t}}, \mathbf{x_{<t}}, \mathbf{c})$ which are parametrized in an autoregressive manner; \ie at each timestep observed and latent variables are conditioned on the past latent samples and observed frames. VRNN therefore uses a \textit{learned prior}~\cite{vrnn, svg}. %
Taking into account the temporal structure of the data, the probability $p(\mathbf{x}, \mathbf{z} | \mathbf{c})$ is factorized as 
\begin{equation}
p(\mathbf{x}, \mathbf{z} | \mathbf{c}) = \prod_{t=1}^T p(x_t | \mathbf{z_{\leq t}}, \mathbf{x_{<t}}, \mathbf{c}) p(z_t| \mathbf{z_{<t}}, \mathbf{x_{<t}}, \mathbf{c}).
\end{equation}

Computing $p(\mathbf{x}|\mathbf{c})$ requires marginalizing over the latent variables $\mathbf{z}$, which is computationaly intractable. %
Instead, VRNN relies on Variation Inference~\cite{jordan1999introduction} and defines an \textit{amortized approximate posterior} $q(\mathbf{z}|\mathbf{x}, \mathbf{c})= \prod_{t=1}^T q(z_t|\mathbf{z_{<t}}, \mathbf{x_{\leq t}}, \mathbf{c})$ that approximates the \textit{true posterior} distribution $p(\mathbf{z}|\mathbf{x},\mathbf{c})$. We then can derive the evidence lower bound (ELBO), a lower bound to the marginal log-likelihood $p(\mathbf{x}|\mathbf{c})$:
\begin{multline}
    \log p(\mathbf{x}|\mathbf{c}) \ge \sum_{t=1}^T \mathbb{E}_{q(\mathbf{z_{\leq t}}|\mathbf{x_{\leq t}}, \mathbf{c})}\log p(x_t | \mathbf{z_{\leq t}}, \mathbf{x_{< t}}, \mathbf{c}) \\
- D_{KL}(q(z_t|\mathbf{z_{<t}}, \mathbf{x_{\leq t}}, \mathbf{c}) || p(z_t| \mathbf{z_{<t}}, \mathbf{x_{< t}}, \mathbf{c}))
\end{multline}
VRNN can be optimized to fit the training data by maximizing the ELBO using stochastic backpropagation and the reparameterization trick~\cite{kingma2013auto,rezende2014stochastic}.

\section{Hierarchical VRNN}

We now introduce a hierarchical version of the VRNN model.
At each timestep, we consider $L$ levels of latents variables $\mathbf{z_t}=({z^1_t}, ..., {z^L_t})$. We then further factorize the latent \textit{prior} as 
\begin{equation}
p(\mathbf{z_t}| \mathbf{z_{<t}}, \mathbf{x_{<t}}, \mathbf{c}) = \prod_{l=1}^L p(z_t^l| \mathbf{z_{t}^{<l}}, \mathbf{z_{<t}^{l}}, \mathbf{x_{<t}}, \mathbf{c}).
\label{eq:hier_prior}
\end{equation}
The sampling process of the latent variable $z_t^l$ now depends on the latent variables from previous time steps $\mathbf{z_{<t}^{l}}$ for that level and on the latent variables of the previous levels at the current timestep $\mathbf{z_t^{<l}}$. Similarly, we can write the \textit{approximate posterior} as:
\begin{equation}
q(\mathbf{\mathbf{z_t}}|\mathbf{z_{<t}}, \mathbf{x_{\leq t}}, \mathbf{c}) = \prod_{l=1}^L q(z_t^l| \mathbf{z_{t}^{<l}}, \mathbf{z_{<t}^{l}}, \mathbf{x_{ \leq t}}, \mathbf{c}).
\label{eq:hier_posterior}
\end{equation}
Using eq.~\ref{eq:hier_prior} and eq.\ref{eq:hier_posterior}, we can rewrite the ELBO as
\begin{multline}
    \log p(\mathbf{x}|\mathbf{c}) \ge \sum_{t=1}^T [\mathbb{E}_{q(\mathbf{z_{t}}|\mathbf{z_{<t}}, \mathbf{x_{\leq t}, \mathbf{c}})} \log p(x_t | \mathbf{z_{\leq t}}, \mathbf{x_{< t}}, \mathbf{c}) \\
- \sum_{l=1}^L D_{KL}(q(z_t^l|\mathbf{z_{t}^{<l}}, \mathbf{z_{<t}^{l}}, \mathbf{x_{\leq t}}, \mathbf{c}) || p(z_t^l| \mathbf{z_{t}^{<l}}, \mathbf{z_{<t}^{l}}, \mathbf{x_{< t}}, \mathbf{c}))].
\label{eq:hier_elbo}
\end{multline}
Refer to the Appendix for a full derivation of the ELBO.

\begin{figure}[t]
 \begin{center}
    \includegraphics[trim=10 68 10 27, clip,width=0.40\textwidth]{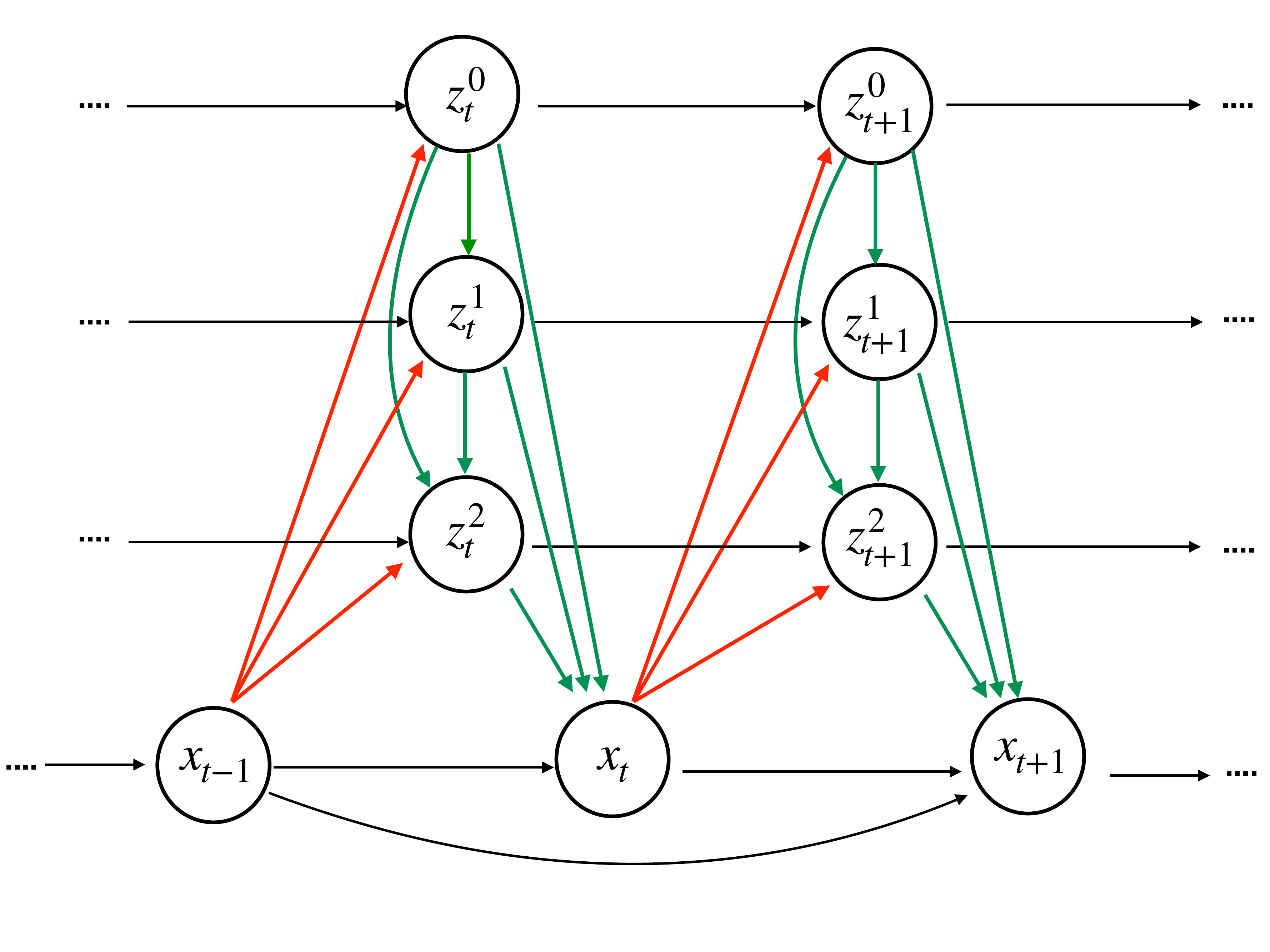}
 \end{center}
    \caption{
    \textbf{Graphical model for the learned prior with the dense latent connectivity pattern}. 
    Arrows in \textcolor{red}{red} show the connections from the input at the previous timestep to current latent variables. 
    Arrows in \textcolor{ForestGreen}{green} highlight skip connections between latent variables and connections to outputs. 
    Arrows in \textbf{black} indicate recurrent temporal connections. We empirically observe that this dense-connectivity pattern eases the training of latent hierarchies.}
    \label{fig:latent_model}
\end{figure}

\begin{figure*}[!t]
    \centering
    \includegraphics[trim=0 15 5 6, clip, width=0.90\textwidth]{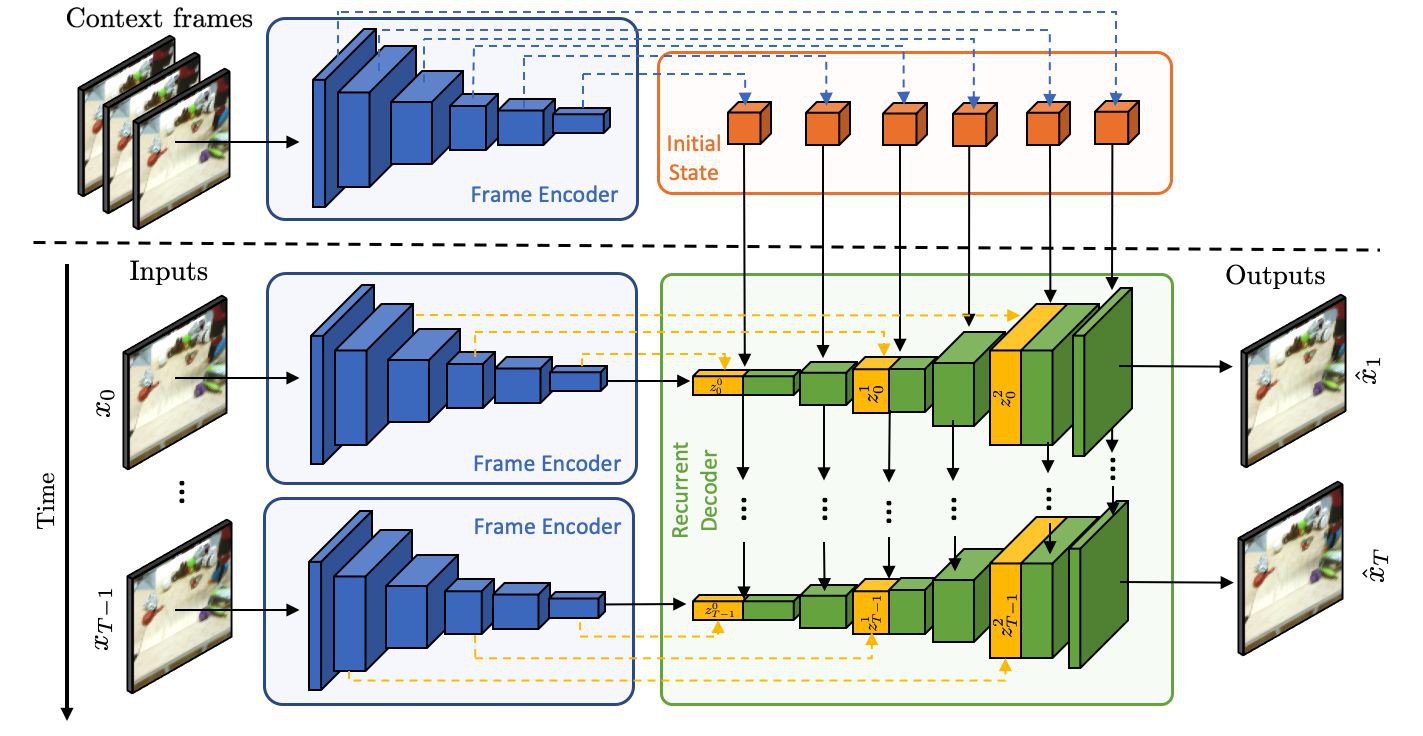}
    \caption{\textbf{Model Parametrization.} Our model uses a CNN to encode frames individually. The representation of the context frames is used to initialize the states of the prior, posterior and likelihood networks, all of which use recurrent networks. At each timestep, the decoder receives an encoding of the previous frame, a set of latent variables (either from the prior or the posterior) and its previous hidden state and predicts the next frame in the sequence.}
    \label{fig:model}
\end{figure*}

\subsection{Dense Latent Connectivity}
\label{sec:denseconnect}

Training a hierarchy of latent variables is known to be challenging as it requires to backpropagate through multiple stochastic layers.
Usually this results in models that only use one specific level of the hierarchy~\cite{kingma2013auto, maaloe2016auxiliary,sonderby2016train}.
To ease the optimization we use a dense connectivity pattern between latent levels both for the prior and the approximate posterior, following~\cite{huang2017densely,maaloe2019biva}.

Fig~\ref{fig:latent_model} illustrates the dense connection of the learned prior (refer to the Appendix for the approximate posterior). For each latent level, the prior and  posterior are  implemented using recurrent neural networks which take as input a  deterministic transformation of $x_{t-1}$ (red arrows in Fig~\ref{fig:latent_model}), and to all the latent variables from the previous levels (green arrows in in Fig~\ref{fig:latent_model}). In addition, each latent variable has a direct connection to the output variables $x_{t}$.

\subsection{Model Parametrization}

We now describe an instantiation of the VRNN model that we will use in the experiments, illustrated in Fig. \ref{fig:model}.
First we compute features for each context frame and use them to initialize the hidden state of the prior/posterior/decoder networks, all of which have recurrent components. At a given timestep $t$, the model takes as input the latent variable samples $\mathbf{z_t}=(z_t^0, ..., z_t^L)$ with the embedding of the previously generated frame $x_{t-1}$ and outputs the next frame $\hat{x}_t$. During training we draw latent samples from the approximate posterior distribution $q(\mathbf{z_t}|\mathbf{z_{< t}}, \mathbf{x_{\leq t}}, \mathbf{c})$ and maximize the ELBO.
To generate unseen sequences, we sample $\mathbf{z_t}$ from the learned prior $p(\mathbf{z_t}|\mathbf{z_{< t}}, \mathbf{x_{< t}}, \mathbf{c})$. Note that since we have multiple levels of conditional latents we use ancestral sampling to generate $\mathbf{z_t}$, \ie we first sample from the top level of the hierarchy and we then sequentially sample the lower levels conditioning on the sampled values of the previous layers in the hierarchy. %

\textbf{Frame Encoder} We use a 2D CNN with ResNet~\cite{he2016deep} blocks and max-pooling layers to represent input frames. 

\textbf{Prior/Approximate Posterior}
We parametrize both the prior and the posterior as a hierarchy of diagonal Normal distributions $\mathcal{N}(\mu, \sigma)$, where the parameters $\mu$ and $\sigma$ are recurrent functions of samples from i) previous levels in the hierarchy and ii) the frame encoder features. Each level in the hierarchy operates at a different spatial resolution, with the top level features operating at a 1x1 resolution, \ie not having a spatial topology. At a given timestep $t$, the parameters for a specific latent level $z^l_t$ are given by a ConvLSTM that consumes i) a previous hidden state,  ii) samples from the previous levels in the hierarchy  $z_t^{<l}$, iii) the feature map of a frame with the same spatial resolution as the ConvLSTM. For the prior network, the input frame embedding corresponds to the previously generated frame $x_{t-1}$, while for the posterior the input comes from the frame to generate $x_t$.

\textbf{Likelihood/Frame Decoder} At each timestep $t$, the decoder takes a representation of the previously generated frame $x_{t-1}$ and the samples $\mathbf{z_t}=(z_t^1, ..., z_t^L)$ and generates $x_t$ according to $p(x_t|\mathbf{z_t}, \mathbf{x_{<t}}, \mathbf{c})$. The decoder consists of ConvLSTMs interleaved with transposed convolutions that upscale the feature maps back to the input resolution. 

\textbf{Initial State} %
The initial states of our prior, posterior and decoder/likelihood models are functions of the context. We use small CNNs to initialize each of the ConvLSTMs layers used in the VAE components. %

\section{Experiments}

All our models are trained with Adam~\cite{kingma2014adam} and a batch size of $b = 128$ on Nvidia DGX-1s. We use a learning rate warmup \cite{goyal2017accurate} starting with an initial learning rate $\lambda$ = 2e-5 that is linearly increased at each timestep until reaching $\lambda$ = 1.6e-4 in 5 epochs. We use $\beta_1$ = 0.5 and $\beta_2$ = 0.9 and weight decay $\delta$ = 1e-4. We train the autoregressive components of our models using teacher forcing~\cite{williams1989learning}. 

Our models are also trained using beta warmup~\cite{sonderby2016ladder}, which consists in gradually increasing the weight of the KL divergence in the ELBO, turning the model from an unregularized Autoencoder into a VAE progressively.
VAEs trained with beta warmup usually encode more information in the latent variables.
Refer to the Appendix for a complete description of our models.

\subsection{Ablation Study}
\label{sec:ablation}

We first investigate the importance of each VRNN component, namely the likelihood, the prior and the posterior.
We focus on the BAIR Push dataset~\cite{dataset_bair_pushing} with 64x64 color sequences of a robotic arm interacting with children toys in a sandbox. Similarly to previous works~\cite{savp}, we use trajectories 256 to 511 as our test set and the rest for training, resulting in the 43264 train and 256 test sequences. At training we randomly subsample 12 frames from each train sequence, use the first 2 frames as the context, and learn to predict the remaining 10 frames.
To evaluate the different model variations, we report the training objective (ELBO) obtained for the training set and the test set.

\subsubsection{Scaling the Likelihood Model}
\label{sec:ablation_likelihood}

\begin{table}[t]
    \small
    \centering
    \begin{tabular}{ccc}
        \toprule
        \textsc{Model} & \textsc{Parameters} & \textsc{Train/Test ELBO}($\downarrow$)\\
        \midrule
        1-ConvLSTM    & 62.22M & 3237/3826 \\
        3-ConvLSTM    & 86.64M  & 1948/2355 \\
        6-ConvLSTM    & 93.81M  & 1279.21/1731.31 \\
        \ \ + higher capacity & 194.15M & 1113.31/1589.72 \\
        \bottomrule
    \end{tabular}
    \caption{\textbf{Ablation - Likelihood} We compare models with different number of recurrent layers for the likelihood network. We observe that the model performance increases monotonically as we add more ConvLSTMs. We further increase the size of the recurrent hidden states for the 6-ConvLSTM model (+ higher capacity variant), also leading to a better data fit. These results suggest that current video prediction models might underfit the data because of reduced likelihood capacity.}
    \label{tab:ablation_capacity}
\end{table}

We assess the importance of the likelihood model $p(x_t | \mathbf{z_{\leq t}}, \mathbf{x_{<t}}, \mathbf{c})$. For this purpose, we build a VRNN with a single level of latent variables and modify the number of ConvLSTM layers in the decoder. 
Our aim is to investigate whether increasing the capacity of the mapping from latent to the observations results in better predictions.

In this experiment, our baseline likelihood model has one LSTM at 1x1 spatial resolution. We then gradually replace convolutional layers in the decoder with ConvLSTM layers, which increases the amount of information that can be carried from previous timesteps and, by extension, also increases the overall likelihood model capacity. We compare to a model with 3 ConvLSTM layers at resolutions 1x1, 4x4 and 8x8 and a model with 6 ConvLSTM layers at 1x1, 4x4, 8x8, 16x16, 32x32 and 64x64. Additionally, we also increase the size of the ConvLSTM layers for the model with 6 layers as another way of adding capacity.

Results can be found in Fig~\ref{tab:ablation_capacity}. We observe that, as a general trend, both the training and test ELBO decrease as we increase the model capacity, which suggests that current video prediction models might operate in an underfitting regime and benefit from higher capacity decoders.

\subsubsection{More Flexible Prior and Posterior}

We now investigate the importance of having more flexible prior and approximate posterior distributions and augment the 6-ConvLSTM VRNN model with a hierarchy of latent variables. 
For all models, we fix the frame encoder and likelihood model\footnote{To add the multiple levels of latents in the decoder we need to modify the likelihood network and slightly increase the number of parameters. However, most ($>85\%$) of the added capacity when adding a new level of latents goes towards the prior and posterior networks.} and change the networks that estimate the learned prior $p(\mathbf{\mathbf{z_t}}|\mathbf{z_{<t}}, \mathbf{x_{<t}}, \mathbf{c})$, and the approximate posterior $q(\mathbf{\mathbf{z_t}}|\mathbf{z_{<t}}, \mathbf{x_{\leq t}}, \mathbf{c})$   over the latent variables. All these models use a dense connectivity pattern and beta warmup.

\begin{table}[t]
  \small
    \centering
    \begin{tabular}{ccc}
        \toprule
        \textsc{Model} & \textsc{Parameters} & \textsc{Train/Test ELBO} ($\downarrow$) \\
        \midrule
        1            & 166.55M & 1141.85/1536.93 \\
        1-8          & 220.60M & 989.39/1313.02 \\
        1-8-32       & 230.74M & \textbf{883.10}/\textbf{1162.24} \\
        1-8-16-32    & 245.19M & 956.63/1256.22 \\
        \midrule
        Naive Training    &  224.18M &  1127.33/1440.58 \\
        BW       &  224.18M &  1101.39/1440.62\\
        Dense    &  230.74M &  1182.60/1547.05 \\
        BW and Dense        &  230.74M &  \textbf{883.10}/\textbf{1162.24} \\
        \bottomrule
    \end{tabular}
    \caption{\textbf{Ablation - Hierarchy of Latents} \textit{Top half:} We compare a VRNN baseline with a single level of latents with no spatial topology (1), a model with two levels of latents at resolutions 1x1 and 8x8 (1-8), our full model with three levels of latents at 1x1, 8x8 and 32x32 (1-8-32), and a model with 4 levels of latents (1-8-16-32). Adding more levels of latents leads to a better fit, with reduced ELBOs. However, adding extra levels of latents without increasing the spatial resolution reduces the performance of the model due to the difficulties in training hierarchical latent variable models. \textit{Bottom half:} To highlight the difficulties in training hierarchies of latents, we investigate the effects of using beta warmup (BW)~\protect\cite{sonderby2016ladder} and having a dense connectivity (Dense) between latents when training the 1-8-32 model. Without these techniques the hierarchy of latents does not bring any benefit compared to the VRNN with 1 level of latent.}
    \label{tab:latent_hierarchy}
\end{table}

We compare a VRNN baseline with a single level of latents with no spatial topology, with a model with two levels of latents at resolutions 1x1 and 8x8 (1-8), three levels of latents at 1x1, 8x8 and 32x32 (1-8-32), and four levels of latents (1-8-16-32) in the top half of Table~\ref{tab:latent_hierarchy}. 
All models are trained with beta warmup and dense latent connectivity.
We observe that in general adding more levels of latents with higher resolution reduces the train and test ELBOs, supporting the hypothesis that a more flexible prior and posterior leads to a better data fit. 
However, we observe diminishing returns past 3 levels, as our 1-8-16-32 model does not outperform the 3 layers model. 
We attribute this to the difficulties in training deep hierarchies of latents, which remains a challenging optimization problem.

To further highlight the difficulties in training hierarchies of latents, we investigate the importance of using beta warmup~\cite{sonderby2016ladder} and having a dense connectivity between latents. 
The results of this experiment can be found in the bottom half of Table~\ref{tab:latent_hierarchy}. We observe that these techniques are required to make our 1-8-32 model make use of the hierarchy of latents and improve upon the single level model.

\begin{figure}[!t]
    \centering
    \includegraphics[trim=5 7 5 5, clip, width=0.98\linewidth]{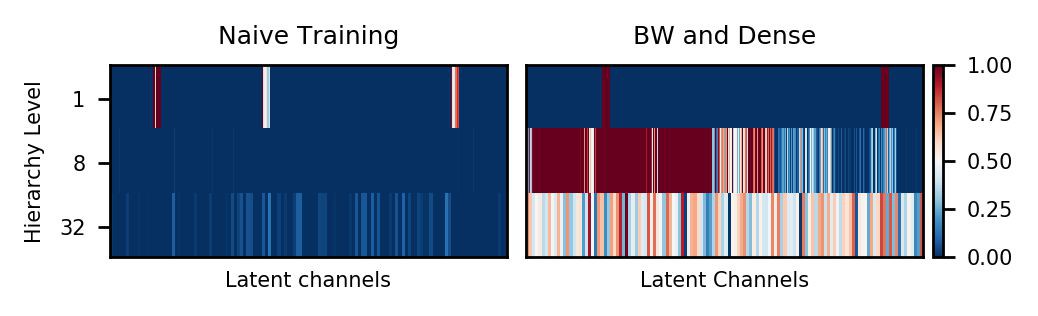}
    \caption{
    \textbf{Average normalized KL per latent channel.} 
    We visualize the mean normalized KL for each latent channel for models from Table~\ref{tab:latent_hierarchy}. 
    Without beta warmup and dense connectivity the hierarchy of latents is underutilized, with most information being encoded in a few latents of the top level. 
    In contrast, the same model with these techniques utilizes all latent levels.}
    \label{fig:kl}
\end{figure}

This is analyzed in more detail in Fig~\ref{fig:kl}, where we visualize the KL between the prior and the posterior distributions for the test sequences of the BAIR Push dataset for the 1-8-32 model and the variant without warmup or dense connectivity (Naive training). 
We consider a channel to be active if its average KL is higher than 0.01 following~\cite{maaloe2019biva}, and consider that a unit with a KL higher than 0.15 is maximally activated. 
We observe that without these techniques the model only uses a few latents of the top level in the hierarchy. 
However, when using beta warmup and a dense connectivity most of the latents are active across levels.

\subsection{Comparisons to Previous Approaches}
Next, we compare our single latent level VRNN (Ours w/o Hier), and our hierarchical VRNN with 3 levels of latents (Ours w/ Hier) to previous video approaches on Stochastic Moving MNIST ~\cite{svg}, BAIR Push \cite{dataset_bair_pushing} and the Cityscapes ~\cite{dataset_cityscapes} datasets.

\subsubsection{Evaluation and Metrics}
\label{sec:evaluation}

Defining evaluation metrics for video prediction is an open research question. %
In general we want models to predict sequences that are \textit{plausible}, look \textit{realistic} and \textit{cover} all possible outcomes.
Unfortunately, we are not aware of any metric that reflects all these aspects.

To measure \textit{coverage} and \textit{plausibility} we adopt the evaluation protocol proposed in~\cite{svg, savp}.
For each ground truth test sequence,
we sample $N$ random predictions from the model which are conditioned on the test sequence initial frames. 
Then we find the sample that best matches the ground truth sequence according to a given metric and report that metric value. 
Some common metric choices are Mean-Square Error (MSE), Structural Similarity (SSIM)~\cite{ssim} or Peak Signal-to-Noise Ratio (PSNR). 
In practice, these metrics have been shown to not correlate well with human judgement as they tend to prefer blurry predictions over sharper but imperfect generations~\cite{lpips, savp, fvd}.
LPIPS~\cite{lpips}, on the other hand, is a perceptual metric that employs CNN features and has better correlation to human judgment.
For this evaluation we choose to produce $N =$ 100 samples following previous work and use SSIM and LPIPS as metrics. 
We have empirically observed that using 100 samples the results stay fairly consistent across different samplings. 
We report the metric average over the test set.

Additionally, we also use the recently proposed Fr\'echet Video Distance (FVD), which measures sample \textit{realism}. 
FVD uses features from a 3D CNN and has also been shown to correlate well with human perception~\cite{fvd}.
FVD compares populations of samples to assert whether they were both generated by the same distribution (it does not compare pairs of ground truth/generated frames directly). 
We form the ground truth population by using all the test sequences with their context. For the predicted population we randomly sample one video out of the $N$ generated for each test sequence. 
We repeat this process 5 times and report the mean of the FVD scores obtained, which stay fairly stable across samplings.

\begingroup
\setlength{\tabcolsep}{3pt} %
\begin{table}[t]
    \small
    \centering
    \begin{tabular}{cccc}
        \toprule
        \textsc{Model} & \textsc{FVD} ($\downarrow$) & \textsc{LPIPS} ($\downarrow$) & \textsc{SSIM} ($\uparrow$) \\
        \midrule
        \textsc{SVG-LP ~\cite{svg}}       & 90.81  & 0.153 $\pm$ 0.03     & 0.668 $\pm$ 0.04\\
        \hdashline
        \textsc{Ours w/o hier}                     & 63.81  & \textbf{0.102} $\pm$ \textbf{0.04} & \textbf{0.763} $\pm$ \textbf{0.09}   \\
        \textsc{Ours w/ hier}  & \textbf{57.17} & \textbf{0.103} $\pm$ \textbf{0.03} & \textbf{0.760} $\pm$ \textbf{0.08}\\
        \bottomrule
    \end{tabular}
    \caption{\textbf{Stochastic Moving MNIST.} We compute the FVD metric between samples from different models and test sequences as well as the average LPIPS and SSIM of the best sample for each test sequence. Our models outperform the SVG-LP baseline on all metrics by a significant margin. While our model with hierarchical latent variables obtains a better FVD score, both variants obtain comparable results in this relatively simple dataset.}
    \label{tab:mnist_metrics}
\end{table}
\endgroup

\begingroup
\setlength{\tabcolsep}{0.5pt} %
\renewcommand{\arraystretch}{1} %
\begin{figure*}[!t]
    \small
    \centering
    \begin{tabular}{ccc|ccccccccccc}
            &
            \multicolumn{2}{c|}{\textbf{Context}}  &
            \multicolumn{11}{|c}{\textbf{Predicted Frames}} 
            \\
            &
            $t = 1$ &
            $t = 2$ &
            $t = 3$ & 
            $t = 4$ & 
            $t = 6$ & 
            $t = 8$ &
            $t = 10$ & 
            $t = 12$ & 
            $t = 15$ & 
            $t = 18$ &
            $t = 20$ &
            $t = 25$ \\
            \midrule
            {\rotatebox[origin=c]{90}{GT}} &
            \raisebox{-.4\height}{\includegraphics[scale=0.60]{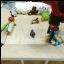}} &
            \raisebox{-.4\height}{\includegraphics[scale=0.60]{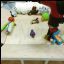}} &
            \raisebox{-.4\height}{\includegraphics[scale=0.60]{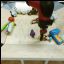}} &
            \raisebox{-.4\height}{\includegraphics[scale=0.60]{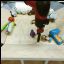}} &
            \raisebox{-.4\height}{\includegraphics[scale=0.60]{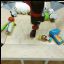}} &           
            \raisebox{-.4\height}{\includegraphics[scale=0.60]{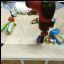}} &            
            \raisebox{-.4\height}{\includegraphics[scale=0.60]{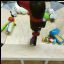}} &
            \raisebox{-.4\height}{\includegraphics[scale=0.60]{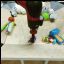}} &
            \raisebox{-.4\height}{\includegraphics[scale=0.60]{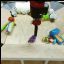}} &
            \raisebox{-.4\height}{\includegraphics[scale=0.60]{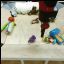}} &
            \raisebox{-.4\height}{\includegraphics[scale=0.60]{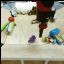}} &
            \raisebox{-.4\height}{\includegraphics[scale=0.60]{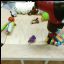}}
            \\[12.5pt]
            \multicolumn{3}{c|}{\textsc{SAVP~\cite{savp}}} &
            \raisebox{-.4\height}{\includegraphics[scale=0.60]{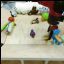}} &
            \raisebox{-.4\height}{\includegraphics[scale=0.60]{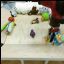}} &
            \raisebox{-.4\height}{\includegraphics[scale=0.60]{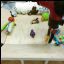}} & 
            \raisebox{-.4\height}{\includegraphics[scale=0.60]{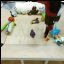}} &            
            \raisebox{-.4\height}{\includegraphics[scale=0.60]{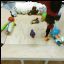}} &
            \raisebox{-.4\height}{\includegraphics[scale=0.60]{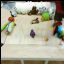}} &
            \raisebox{-.4\height}{\includegraphics[scale=0.60]{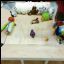}} &
            \raisebox{-.4\height}{\includegraphics[scale=0.60]{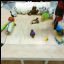}} &
            \raisebox{-.4\height}{\includegraphics[scale=0.60]{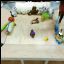}} &
            \raisebox{-.4\height}{\includegraphics[scale=0.60]{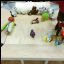}}
            \\[12.5pt]
            \multicolumn{3}{c|}{\textsc{SVG-LP~\cite{svg}}} &
            \raisebox{-.4\height}{\includegraphics[scale=0.60]{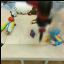}} &
            \raisebox{-.4\height}{\includegraphics[scale=0.60]{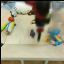}} &           
            \raisebox{-.4\height}{\includegraphics[scale=0.60]{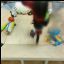}} &           
            \raisebox{-.4\height}{\includegraphics[scale=0.60]{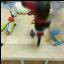}} &            
            \raisebox{-.4\height}{\includegraphics[scale=0.60]{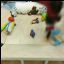}} &
            \raisebox{-.4\height}{\includegraphics[scale=0.60]{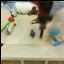}} &
            \raisebox{-.4\height}{\includegraphics[scale=0.60]{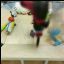}} &
            \raisebox{-.4\height}{\includegraphics[scale=0.60]{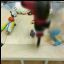}} &
            \raisebox{-.4\height}{\includegraphics[scale=0.60]{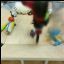}} &
            \raisebox{-.4\height}{\includegraphics[scale=0.60]{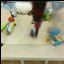}} 
            \\[12.5pt]
            \multicolumn{3}{c|}{\textsc{Ours w/ Hier}} &
            \raisebox{-.4\height}{\includegraphics[scale=0.60]{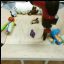}} &
            \raisebox{-.4\height}{\includegraphics[scale=0.60]{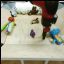}} &           
            \raisebox{-.4\height}{\includegraphics[scale=0.60]{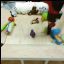}} &           
            \raisebox{-.4\height}{\includegraphics[scale=0.60]{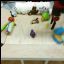}} &
            \raisebox{-.4\height}{\includegraphics[scale=0.60]{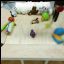}} &
            \raisebox{-.4\height}{\includegraphics[scale=0.60]{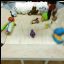}} &
            \raisebox{-.4\height}{\includegraphics[scale=0.60]{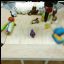}} &
            \raisebox{-.4\height}{\includegraphics[scale=0.60]{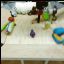}} &
            \raisebox{-.4\height}{\includegraphics[scale=0.60]{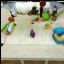}} &
            \raisebox{-.4\height}{\includegraphics[scale=0.60]{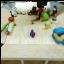}} 
            \\
            \midrule
            {\rotatebox[origin=c]{90}{GT}} &
            \raisebox{-.4\height}{\includegraphics[scale=0.30]{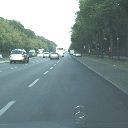}} &
            \raisebox{-.4\height}{\includegraphics[scale=0.30]{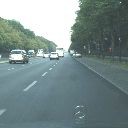}} &
            \raisebox{-.4\height}{\includegraphics[scale=0.30]{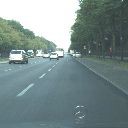}} &
            \raisebox{-.4\height}{\includegraphics[scale=0.30]{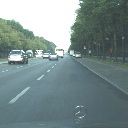}} &           
            \raisebox{-.4\height}{\includegraphics[scale=0.30]{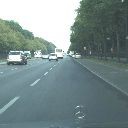}} &
            \raisebox{-.4\height}{\includegraphics[scale=0.30]{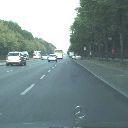}} & 
            \raisebox{-.4\height}{\includegraphics[scale=0.30]{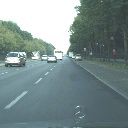}} &
            \raisebox{-.4\height}{\includegraphics[scale=0.30]{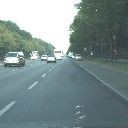}} &
            \raisebox{-.4\height}{\includegraphics[scale=0.30]{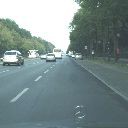}} &
            \raisebox{-.4\height}{\includegraphics[scale=0.30]{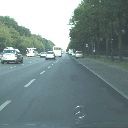}} &
            \raisebox{-.4\height}{\includegraphics[scale=0.30]{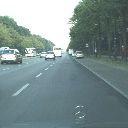}} &
            \raisebox{-.4\height}{\includegraphics[scale=0.30]{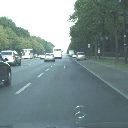}}
            \\[12.5pt]
            \multicolumn{3}{c|}{\textsc{SVG-LP~\cite{svg}}} &
            \raisebox{-.4\height}{\includegraphics[scale=0.30]{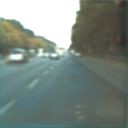}} &
            \raisebox{-.4\height}{\includegraphics[scale=0.30]{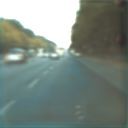}} &
            \raisebox{-.4\height}{\includegraphics[scale=0.30]{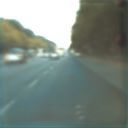}} &
            \raisebox{-.4\height}{\includegraphics[scale=0.30]{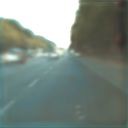}} & 
            \raisebox{-.4\height}{\includegraphics[scale=0.30]{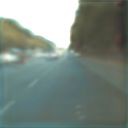}} &
            \raisebox{-.4\height}{\includegraphics[scale=0.30]{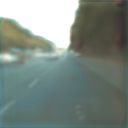}} &
            \raisebox{-.4\height}{\includegraphics[scale=0.30]{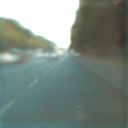}} &
            \raisebox{-.4\height}{\includegraphics[scale=0.30]{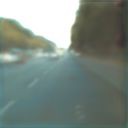}} &
            \raisebox{-.4\height}{\includegraphics[scale=0.30]{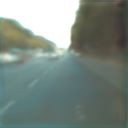}} &
            \raisebox{-.4\height}{\includegraphics[scale=0.30]{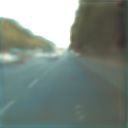}} 
            \\[12.5pt]
            \multicolumn{3}{c|}{\textsc{Ours w/ Hier}} &
            \raisebox{-.4\height}{\includegraphics[scale=0.30]{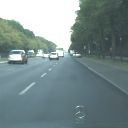}} &
            \raisebox{-.4\height}{\includegraphics[scale=0.30]{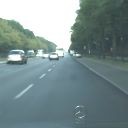}} &          
            \raisebox{-.4\height}{\includegraphics[scale=0.30]{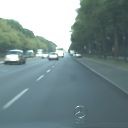}} &
            \raisebox{-.4\height}{\includegraphics[scale=0.30]{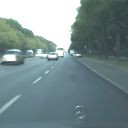}} &
            \raisebox{-.4\height}{\includegraphics[scale=0.30]{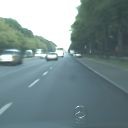}} &
            \raisebox{-.4\height}{\includegraphics[scale=0.30]{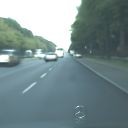}} &
            \raisebox{-.4\height}{\includegraphics[scale=0.30]{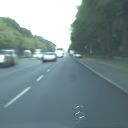}} &
            \raisebox{-.4\height}{\includegraphics[scale=0.30]{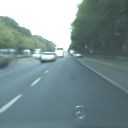}} &
            \raisebox{-.4\height}{\includegraphics[scale=0.30]{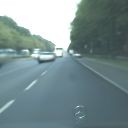}} &
            \raisebox{-.4\height}{\includegraphics[scale=0.30]{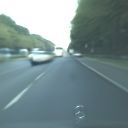}} 

    \end{tabular}
    \caption{\textbf{Selected Samples for BAIR Push and Cityscapes.} We show a sequence for BAIR Push and Cityscapes and random generations from our model and baselines. On BAIR Push we observe that the SAVP predictions are crisp but sometimes depict inconsistent arm-object interactions. SVG-LP produces blurry predictions in uncertain areas such as occluded parts of the background or those showing object interactions. Our model generates plausible interactions with reduced blurriness relatively to SVG-LP. On Cityscapes, the SVG-LP baseline is unable to model any motion. Our model, using a hierarchy of latents, generates more visually compelling predictions. More samples can be found in the Appendix.}
    \label{fig:samples_pushbair}
\end{figure*}
\endgroup

\subsubsection{Stochastic Moving MNIST}

Stochastic Moving MNIST is a synthetic dataset proposed in~\cite{svg} which consists of black and white sequences of MNIST digits moving over a black background and bouncing off the frame borders. As opposed to the original Moving MNIST dataset~\cite{video_lstm} with completely deterministic motion, Stochastic Moving MNIST has uncertain digit trajectories - the digits bounce off the border with a random new trajectory. %
We train two variants of our model and compare to the SVG-LP baseline~\cite{svg}, for which we use a pretrained model from the official codebase. 
All models are trained using 5 frames of context and 10 future frames to predict. To evaluate the models, we follow the procedure in~\cite{svg} described in section~\ref{sec:evaluation}. 

We report the results of the experiment in Table~\ref{tab:mnist_metrics}. We observe that both versions of our model (with/out the latent hierarchy) outperform the SVG-LP baseline by a significant margin on all metrics. Note that LPIPS and FVD might not be suited to this dataset as they use features from CNNs trained on real world images, but we report them for completeness. 
Visually, our samples (found in the Appendix) reproduce the digits more faithfully with reduced degradation over time. 
There are small differences between the two versions of our model, suggesting that the extra expressiveness of the hierarchical model is not necessary in this synthetic dataset.

\subsubsection{BAIR Push}

\begin{figure*}[!t]
    \centering
    \begin{subfigure}[c]{0.53\textwidth}
        \raisebox{-.5\height}{\includegraphics[trim=30 0 30 10, clip, width=0.88\textwidth]{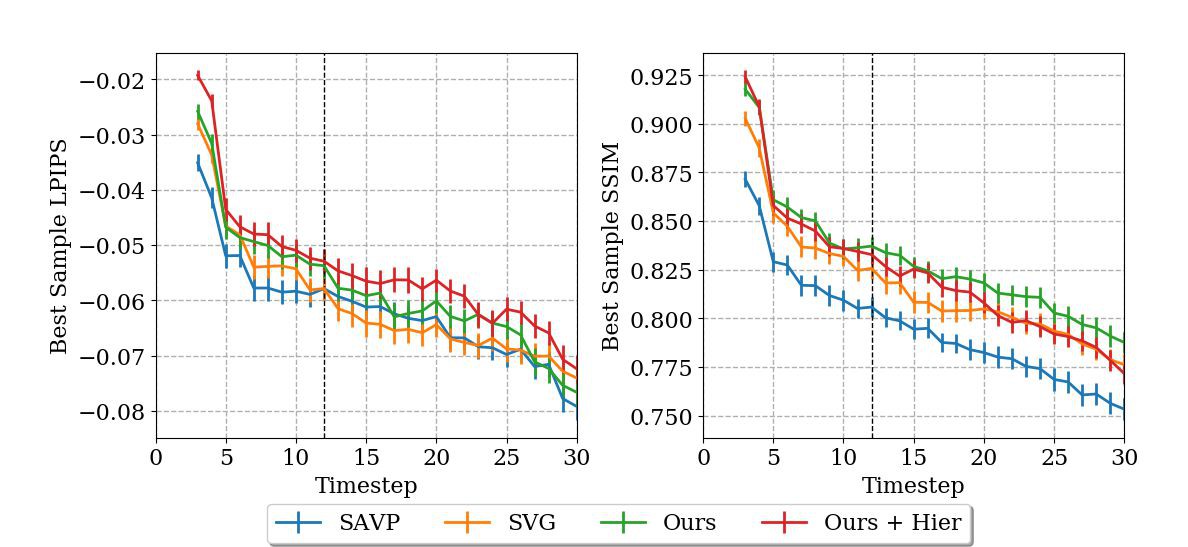}}
    \end{subfigure}\hfill
    \begin{subfigure}[c]{0.47\textwidth}
        \begingroup
    
        \setlength{\tabcolsep}{3pt} %
        \begin{tabular}{cccc}
            \toprule
            \textsc{Model} & \textsc{FVD} ($\downarrow$) & \textsc{LPIPS} ($\downarrow$) & \textsc{SSIM} ($\uparrow$) \\
            \midrule
            \textsc{SVG-LP \cite{svg}}      & 256.62           & $0.061 \pm 0.03$    & $0.816 \pm 0.07$ \\
            \textsc{SAVP \cite{savp}}       & \textbf{143.43}  & $0.062 \pm 0.03$  & $0.795 \pm 0.07$ \\
            \hdashline
            \textsc{Ours w/o Hier}                    & 149.22          & $0.058 \pm 0.03$  & \textbf{0.829} $\pm$ \textbf{0.06} \\
            \textsc{Ours w/ Hier}  & \textbf{143.40} & \textbf{0.055} $\pm$ \textbf{0.03} & $0.822 \pm 0.06$ \\
            \bottomrule
        \end{tabular}
        \endgroup
    \end{subfigure}
    \caption{\textbf{BAIR Push - Results}. \textit{Left:} We show the evolution in time of the Average LPIPS and SSIM of the best predicted sample per test sequence. \textit{Right:} We report the Average FVD, SSIM and LPIPS of the best sample for each test sequence. Compared to SVG-LP, both our model with a single level of latents and the hierarchical models improve all metrics. Compared to SAVP, we obtain better LPIPS and SSIM. Our model with a single level of latents performs better in SSIM but worse on perceptual metrics. When adding the hierarchy of latents, our model matches the FVD of SAVP and improves the LPIPS, indicating samples of similar visual quality and better coverage of the ground-truth sequences.}
    \label{fig:metrics_pushbair}
\end{figure*}

\begin{figure*}[!t]
    \centering
    \begin{subfigure}[c]{0.53\textwidth}
        \raisebox{-.5\height}{\includegraphics[trim=30 0 30 10, clip, width=0.88\textwidth]{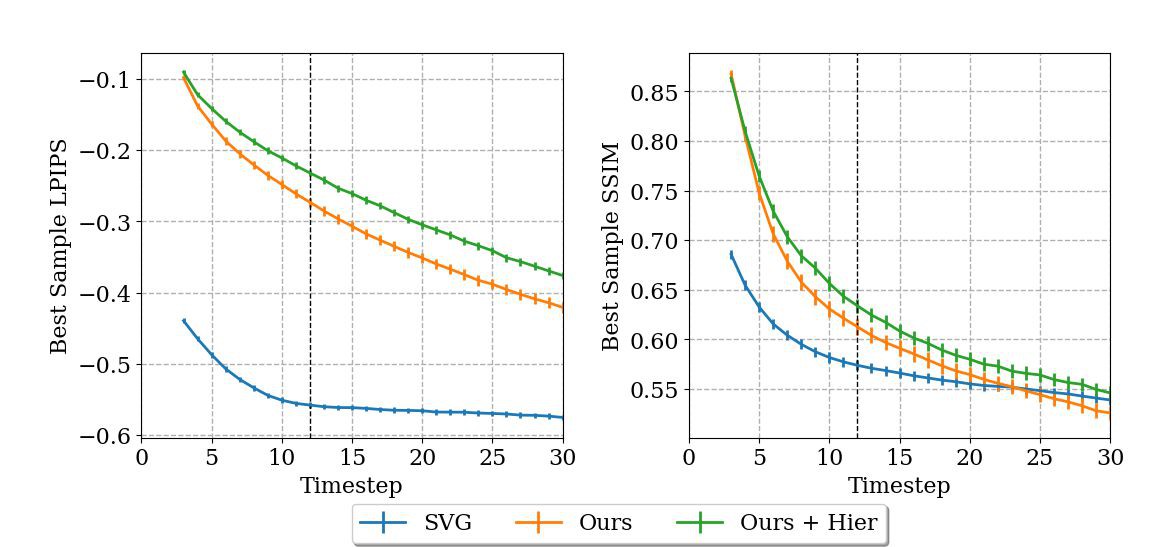}}
    \end{subfigure}\hfill
    \begin{subfigure}[c]{0.47\textwidth}
        \begingroup
        \setlength{\tabcolsep}{3pt} %
        \begin{tabular}{cccc}
            \toprule
            \textsc{Model} & \textsc{FVD} ($\downarrow$) & \textsc{LPIPS} ($\downarrow$) & \textsc{SSIM} ($\uparrow$) \\
            \midrule
            \textsc{SVG-LP ~\cite{svg}}       & 1300.26     & $0.549 \pm 0.06$     & $0.574 \pm 0.08$ \\
            \hdashline
            \textsc{Ours w/o Hier}                     & 682.08      & $0.304 \pm 0.10$     & $0.609 \pm 0.11$ \\
            \textsc{Ours w/ Hier}  & \textbf{567.51} & \textbf{0.264} $\pm$ \textbf{0.07} & \textbf{0.628} $\pm$ \textbf{0.10} \\
            \bottomrule
        \end{tabular}
        \endgroup
    \end{subfigure}
    \caption{\textbf{Cityscapes - Quantitative Results} We report FVD, SSIM and LPIPS scores on Cityscapes at 128x128 resolution for the SVG-LP ~\protect\cite{svg} baseline and two variants of our model. Increasing the capacity of the likelihood model brings an improvement in all metrics over the SVG baseline.
    When adding a hierarchy of latents we observe further improvements, validating its usefulness. Even though SVG matches our models in SSIM at later timesteps, this does not correlate well with human judgement, as the generated SVG samples show more blurriness (see Fig.~\ref{fig:samples_pushbair}).}
    \label{fig:metrics_cityscapes}
\end{figure*}

We compare our VRNN models to SVG-LP~\cite{svg} and SAVP~\cite{savp}. We use their official implementations and pretrained models to reproduce their results.
We use the experimental setup of previous works ~\cite{svg, savp}, using 2 context frames and generating 28 frames.%

Results can be found on  Fig~\ref{fig:metrics_pushbair}. When the robotic arm is interacting with an object, SVG-LP tends to generate blurry predictions characterized by a high FVD score.  SAVP exhibits a lower FVD as it produces more realistically looking predictions. However, SAVP does not have a better coverage of the ground truth sequences compared to SVG-LP as measured by LPIPS and SSIM. By inspecting the SAVP samples we notice that the SAVP generations tend to be sharper but sometimes they exhibit temporal inconsistencies or implausible interactions (see Fig~\ref{fig:samples_pushbair}). 
Our w/o Hier VRNN models obtain better scores than SVG-LP, the previous best VAE-type model. This supports the importance of having a high-capacity likelihood model. In addition, our hierarchical VRNN further improves both the FVD and LPIPS metrics, suggesting that the hierarchy of latents helps modeling the data
In particular, our hierarchical VRNN shows an improvement of $44\%$ in terms of FVD and $9.8\%$ in terms of LPIPS over SVG-LP, the previous best VAE-based model. It also achieves a similar FVD than the SAVP GAN-VAE model, while outperforming it in terms of LPIPS by  $11.2\%$.

\subsubsection{Cityscapes}

The Cityscapes dataset contains sequences recorded from a car driving around multiple cities under different conditions. Cityscapes is a challenging dataset - while contiguous frames are locally similar, uncertainty grows significantly with time. Compared to previous datasets, the backgrounds in Cityscapes do not stay constant across time.

We consider sequences with 30 frames from the training set cities for a total of 1877 train sequences and randomly select 256 test sequences. We use 2 context and 10 prediction frames to train the models. 
At test time we predict 28 frames following the BAIR Push experimental protocol. 
We preprocess the videos by taking a 1024x1024 center crop of the original sequences and resizing them to 128x128 pixels. For evaluating the models we use the standard setup where we generate 100 samples per test sequence and report FVD, SSIM and LPIPS metrics. Since none of the baselines from previous experiments are trained on Cityscapes, we use the official SVG implementation (that defines models with 128x128 inputs) and train a SVG-LP model. We train all models for 100 epochs.

Results for this experiment can be found in Fig.~\ref{fig:metrics_cityscapes}. 
SVG-LP has trouble modelling motion in the dataset, usually predicting a static image similar to the last context frame. 
In contrast, our model without a hierarchy of latents is able to model the changing scene. When adding hierarchical latents our model is able to capture more fine-grained details, and as a result, it produces more visually appealing samples with a boost in all metrics. 
We note that the SSIM scores for SVG-LP match those of our models at later timesteps in the prediction, however this does not translate to better samples as can be seen in Fig.~\ref{fig:samples_pushbair} or in the Appendix.
This further indicates that SSIM might not be suitable to evaluate video prediction models.

\section{Conclusions}
We propose a hierarchical VRNN for video prediction that features an improved likelihood model and a hierarchy of latent variables. 
Our approach compares favorably to current state of the art models in terms of the Fr\'echet Video Distance, LPIPS and SSIM metrics, producing visually appealing and coherent samples. Our results demonstrate that current video prediction models benefit from increased capacity, opening the door to further gains with bigger and more flexible generative models.

\clearpage
{\small

}

\end{document}